\documentclass[pdflatex,sn-mathphys-num]{sn-jnl}

\usepackage{graphicx}%
\usepackage{multirow}%
\usepackage{amsmath,amssymb,amsfonts}%
\usepackage{amsthm}%
\usepackage{mathrsfs}%
\usepackage{anyfontsize}
\usepackage[title]{appendix}%
\usepackage{xcolor}%
\usepackage{textcomp}%
\usepackage{manyfoot}%
\usepackage{booktabs}%
\usepackage{algorithm}%
\usepackage{algorithmicx}%
\usepackage{algpseudocode}%
\usepackage{listings}%
\usepackage{hyperref}%
\usepackage{float}%
\usepackage{adjustbox}%
\usepackage{amsmath}
\usepackage{amsbsy}  
\usepackage{fix-cm}
\usepackage{lmodern}

\theoremstyle{thmstyleone}%

\theoremstyle{thmstyletwo}%

\theoremstyle{thmstylethree}%
\newcounter{fignum}
\newcommand{\includecroppedfigure}[1][]{%
    \includegraphics[page=\thefignum,#1]{Figure_cropped.pdf}%
    \stepcounter{fignum}%
}
\begin{document}

\title[PCP-GAN]{PCP-GAN: \textbf{P}roperty-\textbf{C}onstrained \textbf{P}ore-scale image reconstruction via conditional \textbf{G}enerative \textbf{A}dversarial \textbf{N}etworks}

\author[1]{\fnm{Ali} \sur{Sadeghkhani}}

\author[1]{\fnm{Brandon} \sur{Bennett}}

\author[2]{\fnm{Masoud} \sur{Babaei}}

\author*[1]{\fnm{Arash} \sur{Rabbani}}\email{a.rabbani@leeds.ac.uk}

\affil*[1]{\orgdiv{School of Computer Science}, \orgname{University of Leeds}, \orgaddress{\city{Leeds}, \postcode{LS2 9JT}, \country{UK}}}

\affil[2]{\orgdiv{Department of Chemical Engineering}, \orgname{The University of Manchester}, \orgaddress{\city{Manchester}, \postcode{M13 9PL}, \country{UK}}}

\abstract{Obtaining truly representative pore-scale images that match bulk formation properties remains a fundamental challenge in subsurface characterization, as natural spatial heterogeneity causes extracted sub-images to deviate significantly from core-measured values. This challenge is compounded by data scarcity, where physical samples are only available at sparse well locations. This study presents a multi-conditional Generative Adversarial Network (cGAN) framework that generates representative pore-scale images with precisely controlled properties, addressing both the property-matching challenges and data availability constraints. The framework was trained on thin section samples from four depths (1879.50~m to 1943.50~m) of a carbonate formation, simultaneously conditioning on porosity values and depth parameters within a single model. Unlike previous approaches requiring separate models for different formations, the framework processes RGB thin section images that preserve critical mineralogical information (anhydrite-dolomite differentiation, grain boundaries, interparticle-intraparticle porosity distinctions) lost in conventional grayscale or binarized representations. This approach captures both universal pore network principles and depth-specific geological characteristics, from grainstone fabrics with interparticle-intercrystalline porosity to crystalline textures with anhydrite inclusions. The model achieved strong porosity control (R$^2$ = 0.95) across all formations with mean absolute errors of 0.0099--0.0197. Morphological validation confirmed preservation of critical pore network characteristics including average pore radius, specific surface area, and tortuosity, with statistical differences remaining within acceptable geological tolerances, consistent with geological authenticity. Two-point correlation ($S_2$) analysis further confirmed that the generated images preserve the spatial continuity and characteristic length scales of natural pore networks rather than reproducing porosity in a spatially incoherent manner, and both porosity control and reconstruction accuracy remained consistent across the range of imaging resolutions tested (1.8--3.0~$\mu$m/pixel).

When validated against core sample properties, generated images demonstrated higher property fidelity with dual-constraint errors (combined porosity-permeability deviations) of 1.9--12.4\% compared to 37.5--713.6\% for randomly extracted real sub-images, showing tighter clustering around target porosity-permeability values. This capability to generate geologically authentic images that better match bulk formation properties than traditional sampling provides practical tools for subsurface characterization, particularly valuable for modeling processes in carbon storage, geothermal energy, and groundwater management, where knowing the characteristic morphology of the pore space is critical for implementing digital rock physics.}


\maketitle

\section{Introduction}

Subsurface fluid flow analysis constitutes an essential component of geological and engineering applications, spanning carbon storage, groundwater management, geothermal energy development, and contaminant remediation. Central to these applications is digital rock analysis, which provides critical insights into complex pore-scale structures governing fluid transport phenomena \citep{Blunt2013Pore-scaleModelling}, but faces persistent challenges due to visual pore-scale data scarcity despite advanced imaging technologies (micro-CT, FIB-SEM, laser scanning confocal microscopy) being limited by practical constraints \citep{Cnudde2013High-resolutionApplications, Li2018DirectQuantification}.

These data are typically only available at specific depths and wells due to high core extraction costs, technical difficulties, and physical sampling constraints. This scarcity creates significant gaps in our understanding of subsurface formations, particularly when attempting to characterize formations between sampled points. The challenge is further compounded by the heterogeneous nature of porous media, where physical properties can vary significantly within the material, requiring numerous experiments with limited reproducibility \citep{Neuman1990UniversalMedia} and making digital rock analysis and fluid flow simulations particularly difficult.

Traditional approaches to address this data scarcity often rely on interpolation or statistical methods, such as the Joshi-Quiblier-Adler (MJQA) method \citep{Adler1990FlowMedia, Quiblier1984AMedia} and simulated annealing techniques \citep{Yeong1998ReconstructingMedia}. However, these methods may not adequately capture the complex heterogeneity of subsurface formations. For instance, the MJQA method tends to underestimate pore-space connectivity and struggles to replicate the inherent anisotropy of real porous structures \citep{PAL-ERICREN2002ProcessProperties}, while simulated annealing, despite its versatility, is limited by significant computational demands \citep{Alexander2009HierarchicalImages}.

More sophisticated methods are needed to generate representative pore-scale images for depths lacking physical samples. Deep learning methods demonstrate remarkable capabilities in extracting complex data features and establishing intricate mappings between input and output patterns \citep{Goodfellow2016DeepLearning, Lecun2015DeepLearning}, enabling diverse pattern capture within porous images that makes them particularly valuable for geological microstructure reconstruction \citep{Karpatne2019MachineOpportunities, Feng2020AnLearning, Amiri2024NewGANs, Li2022CascadedImage, Huang2022Deep-learning-basedMethod}. 

Among these deep learning approaches, Generative Adversarial Networks (GANs) represent a particularly powerful framework, employing an adversarial process between generator and discriminator networks to capture and replicate complex patterns in training data \citep{Goodfellow2014GenerativeNetworks}. While GANs have shown promise in geological modeling tasks, including porous media reconstruction and rock parameter prediction \citep{Feng2020AnLearning}, most existing studies focus on unconditional generation and do not incorporate specific constraints on attributes that are crucial for accurate subsurface characterization.

Conditional Generative Adversarial Networks (cGANs) extend traditional GAN frameworks by incorporating additional input information, enabling targeted and controllable image synthesis \citep{Mirza2014ConditionalNets}. This conditional approach enables porous media image generation with specific desired properties, offering practical tools for geoscientists and engineers. However, existing applications in subsurface characterization remain limited, particularly regarding detailed validation of generated images' geological realism and quantitative property preservation.

The development of cGANs has seen significant advancement in material science applications, particularly in controlling specific properties during generation. Early implementations focused on basic property control, with researchers showing the ability to condition generation on fundamental characteristics such as phase distribution and volume fractions. Kishimoto et al. \citep{Kishimoto2023ConditionalFractions} developed a conditional framework for generating three-dimensional porous structures of solid oxide fuel cell (SOFC) anodes with controllable volume fractions, achieving the capability to generate realistic microstructures with precise phase distributions, even for compositions outside the training data range.

Building on single-property control, researchers have integrated physical constraints and process parameters into conditional generation frameworks. Yang et al. \citep{Yang2022Multi-scaleNetworks} developed a multi-scale reconstruction method using low-resolution images as conditional input for a cGAN framework, preserving both micro and macro features while maintaining physical accuracy. Tang et al. \citep{Tang2021MachineAlumina} implemented regression-based conditioning using numerical processing parameters as conditions for predicting microstructures during laser sintering, allowing interpolation between trained conditions and enabling microstructure prediction for unexplored processing parameters.

A distinct approach has emerged focusing on statistical property control and hybrid methodologies. Shams et al. \citep{Shams2021AST-CGAN} introduced ST-CGAN, a hybrid approach that combines statistical methods with cGANs for 3D porous media reconstruction from 2D images, where the statistical component provides conditional input to cGAN, allowing more realistic reconstructions of homogeneous and heterogeneous samples while improving network stability and efficiency compared to conventional methods.

The most sophisticated implementations achieve simultaneous control of multiple material characteristics. Zhou and Wu \citep{Zhou20233DLearning} introduced a multi-conditional generative adversarial network (MCGAN) for 3D reconstruction of digital rocks guided by multiple petrophysical parameters, incorporating porosity, specific surface area, fractal dimension, and tortuosity through a switch structure that allows selective parameter enabling. Similarly, Zheng and Zhang \citep{Zheng2022DigitalNetworks} demonstrated simultaneous control over rock type classification, porosity, and correlation length (both isotropic and anisotropic) while leveraging progressive growing architecture for improved stability.

Recent work has explored physics-informed conditioning and alternative generative architectures for porous media reconstruction. Ren and Srinivasan \citep{Ren2024UsingMedia} extended GAN-based approaches by introducing physics-informed post-training optimization, embedding a pore network model within an iterative loop to perturb the latent space of a Wasserstein DCGAN until target properties are achieved. Moving beyond GANs, Naiff et al. \citep{Naiff2026ControlledReconstruction} introduced controlled latent diffusion models within the Elucidated Diffusion Models (EDM) framework for 3D reconstruction, demonstrating that porosity conditioning alone can ensure consistent representation of multiple properties while enabling generation of larger volumes (256$^3$ voxels) than previous diffusion approaches.

Despite these advances in conditional generation, three critical limitations persist in existing geological applications. First, no study has implemented simultaneous control of porosity with depth-specific geological characteristics within an integrated architecture. While Zhou and Wu \citep{Zhou20233DLearning} demonstrated multi-parameter control (porosity, surface area, fractal dimension, tortuosity), their approach focused on a single formation type without addressing depth-dependent geological variability. Similarly, Zheng and Zhang \citep{Zheng2022DigitalNetworks} achieved rock type and porosity conditioning, but required separate models for each rock type classification rather than learning formation-specific characteristics across a continuous depth range within a unified architecture. This limits practical applicability for characterizing heterogeneous subsurface formations where geological properties vary continuously with depth.

Second, and more critically, existing implementations have relied exclusively on grayscale or binary representations, fundamentally losing critical petrographic information essential for accurate geological characterization. RGB thin section images with blue-dyed epoxy preserve three categories of information unrecoverable from simplified representations: (1) mineralogical differentiation between non-porous anhydrite (white) and porous dolomite (gray) matrices, critical for predicting reactive transport behavior; (2) grain boundary identification essential for fabric classification (grainstone versus crystalline), which controls flow architecture \citep{Lucia1995Rock-fabric/petrophysicalCharacterization}; and (3) distinction between interparticle and intraparticle porosity systems that exhibit fundamentally different flow behaviors and connectivity patterns. These color-encoded relationships enable simultaneous visualization of mineralogical composition, diagenetic alterations, and pore architecture that directly influence subsurface fluid transport.

Third, this integration is particularly challenging as it requires the model to learn both generalizable pore geometry relationships and formation-specific features that vary with samples and diagenetic history, while preserving these complex color-encoded geological relationships. Previous work has not demonstrated this capability across diverse carbonate fabrics and mineralogical assemblages.

To address these limitations and advance the field of digital rock physics, this study makes four key contributions. First, we develop the first unified multi-conditional GAN framework that simultaneously learns porosity-depth-pore network relationships across multiple geological formations without requiring formation-specific models. Second, we demonstrate that RGB thin section images enable preservation of critical mineralogical and textural information that fundamentally influences flow predictions but is lost in grayscale approaches. Third, we introduce a dual-constraint validation framework that quantifies not only morphological preservation but also demonstrates that synthetically generated images can better represent bulk formation properties than randomly extracted sub-images from naturally heterogeneous cores. Fourth, we validate the framework across diverse carbonate fabrics spanning grainstones with dual porosity systems to crystalline formations with anhydrite inclusions, confirming geological authenticity through morphological analysis, spatial-continuity ($S_2$) analysis, and robustness testing across imaging resolutions.

This work extends beyond methodological advancement to address a fundamental challenge in digital rock analysis, which is obtaining pore-scale images that match bulk formation properties from naturally heterogeneous formations. By generating representative images for specific porosity values at trained depths, the framework achieves order-of-magnitude improvements in property fidelity, with substantially lower deviations than randomly extracted sub-images. This shows that the framework has learned fundamental pore geometry-petrophysical property relationships rather than merely replicating visual patterns. Producing geologically authentic images with precisely controlled properties that better represent bulk formation characteristics than natural sub-samples directly enhances formation analysis, fluid flow simulation accuracy, and decisions across subsurface applications.

\section{Methodology}
\label{sec:me}

\subsection{Data Description and Preparation}

This section describes the dataset and data preparation process for training the cGAN model. Our goal is to train the cGAN to generate realistic, colour-coded porous medium images with specific characteristics, effectively capturing the complex structure of rock porosity as revealed through specialized imaging techniques. The following subsections detail the sample characteristics and the automated porosity quantification methodology.

\subsubsection{Dataset and Sample Characteristics}

In the current study, the dataset for training the developed cGAN model was prepared from petrography images (Figure \ref{fig:petrography_sample_images}) comprising four distinct sets of thin-section samples from a carbonate formation \citep{Rabbani2017EstimationData}. These samples are designated as Sample 1 (1879.50 m), Sample 2 (1881.90 m), Sample 3 (1918.50 m), and Sample 4 (1943.50 m), where the values in parentheses indicate the sampling depths within the carbonate formation. The thin section images consist of dolomite, limestone, and anhydrite lithology. Carbonate rocks exhibit complex pore networks resulting from various geological processes \citep{Eberli2003FactorsRocks}, providing geologically challenging training data. In these images, the blue areas represent porous regions that result from the blue-dyed epoxy resin injected during the preparation of the sample to differentiate the pore space from rock minerals. The thin section images were available at three resolutions (approximately 1.8, 2.25, and 3.0~$\mu$m/pixel), where the 3.0~$\mu$m/pixel images were used for the main analysis.

Table \ref{tab:sample_characteristics} presents the comprehensive geological and petrophysical characteristics of the four carbonate samples used in this study. The samples exhibit diverse geological features ranging from grainstone to crystalline fabrics, with core porosity values spanning from 10.58\% to 24.77\% and permeability ranging from 12.09 to 181.44 mD. This diversity in pore structure characteristics and lithological properties provides an ideal dataset for training the multi-conditional GAN model across different geological formations within the carbonate sequence.

\begin{table}[htbp]
\caption{Geological and petrophysical characteristics of carbonate samples used for cGAN training.}
\label{tab:sample_characteristics}
\footnotesize
\begin{tabular}{|c|c|c|c|c|c|c|}
    \hline
    \textbf{Sample} & \textbf{Depth} & \textbf{Core} & \textbf{Core} & \textbf{Fabric} & \textbf{Main Pore} & \textbf{Lithology} \\
    \textbf{Number} & \textbf{(m)} & \textbf{Porosity (\%)} & \textbf{Permeability (mD)} & \textbf{Type} & \textbf{Type} & \\
    \hline
    1 & 1879.50 & 15.73 & 33.64 & Grainstone & IP-IT & Dolomite \\
    \hline
    2 & 1881.90 & 24.77 & 181.44 & Grainstone & IP-IT & Dolomite \\
    \hline
    3 & 1918.50 & 10.58 & 13.39 & Crystalline & IT & Dolomite-Anhydrite \\
    \hline
    4 & 1943.50 & 13.32 & 12.09 & Crystalline & IT & Dolomite \\
    \hline
    \multicolumn{7}{|l|}{\footnotesize \textbf{Note:} IP-IT = Interparticle-Intercrystalline; IT = Intercrystalline} \\
    \hline
\end{tabular}
\end{table}

Throughout this manuscript, samples are referenced by their numerical designation (1-4) with geological and depth details available in Table \ref{tab:sample_characteristics}.

\begin{figure}[H]
    \centering
    \includecroppedfigure[width=0.8\textwidth]
    \caption{Examples of thin-section images (3.0~$\mu$m/pixel) from the carbonate formation from different core samples: (a) Sample 1 (1879.50 m), (b) Sample 2 (1881.90 m), (c) Sample 3 (1918.50 m), and (d) Sample 4 (1943.50 m). Blue areas represent porous regions visualized by blue-dyed epoxy resin, with each image demonstrating the characteristic pore structure variability at different depths.}
    \label{fig:petrography_sample_images}
\end{figure}

Prior to model training, Representative Elementary Volume (REV) analysis determined optimal sub-image size balancing statistical representativeness with porosity variability preservation for cGAN training. Sub-image sizes from 64×64 to 516×516 pixels were systematically evaluated using approximately 10 original images (768×516 pixels) per depth to capture intra-sample spatial heterogeneity, with porosity standard deviation analyzed across all samples following established REV methodologies \citep{Costanza-Robinson2011RepresentativeImplications,Zubov2024InExample}.

The analysis revealed a critical trade-off, while larger sub-images achieve improved mean porosity convergence through spatial averaging, they simultaneously reduce porosity variability essential for training conditional GANs that require diverse porosity distributions. To address this trade-off, a porosity standard deviation threshold of $\sigma = 0.06$ was established through systematic convergence analysis across all samples. At $480 \times 480$ pixels, this threshold was achieved across all geological formations (Figure~\ref{fig:rev_analysis}), providing optimal balance between statistical validity and porosity diversity necessary for effective conditional generation.

This selection ensures training patches capture sufficient spatial heterogeneity for pore network characterization. It also enables generating images matched to measured porosity, ensuring statistical convergence while preserving the geological heterogeneity essential for subsurface characterization.

\begin{figure}[H]
    \centering
    \includecroppedfigure[width=0.8\textwidth]
    \caption{Representative Elementary Volume (REV) analysis depicting porosity standard deviation analysis showing variability reduction with increasing sub-image size for four carbonate samples, with the established threshold ($\sigma = 0.06$) for maintaining adequate variability in cGAN training. The selected 480$\times$480 pixel size (red dashed line) optimizes both statistical validity and porosity diversity essential for effective conditional generation.}
    \label{fig:rev_analysis}
\end{figure}

\subsubsection{Automated Porosity Quantification via Enhanced U-Net}
\label{subsec:Enhanced U-Net}

Accurate porosity quantification forms the foundation of our conditional generation framework, requiring precise segmentation to distinguish pore spaces from solid matrix in thin-section images to establish reliable porosity labels for training. To achieve this, we employed an enhanced U-Net architecture, a convolutional neural network originally developed for biomedical image segmentation \citep{Ronneberger2015U-Net:Segmentation}. U-Net's encoder-decoder structure with skip connections preserves fine-grained spatial information while capturing semantic context, making it particularly suitable for pore segmentation where precise boundary delineation is critical. The architecture's symmetric design allows it to work effectively with limited training data by leveraging both contracting (encoder) and expanding (decoder) paths to learn robust feature representations. 

We enhanced the standard U-Net by incorporating attention gates \citep{Oktay2018AttentionPancreas} and deep supervision mechanisms to improve performance on complex carbonate pore structures. The attention gates selectively emphasize relevant features at skip connections, improving boundary delineation in complex pore structures characteristic of carbonate formations where boundaries between pore and solid phases can be indistinct due to partial dolomitization and varying crystal sizes. The network processes 480 × 480 × 3 RGB thin section images to produce corresponding binary masks, where the pore spaces are distinguished from the solid matrix material.

Training utilized a hybrid loss function combining Dice coefficient and binary cross-entropy (BCE) with equal weighting (0.5:0.5), addressing the inherent class imbalance between pore and solid phases while ensuring both accurate region overlap and pixel-wise classification. The segmentation model achieved high performance across the test dataset of 938 samples, with mean Dice coefficient of $0.952 \pm 0.041$, IoU of $0.910 \pm 0.071$, and accuracy exceeding $0.956 \pm 0.032$. Porosity values calculated from the predicted masks demonstrated strong agreement with ground truth measurements, exhibiting mean absolute error of 0.018, confirming the reliability of automated porosity quantification. This pre-trained segmentation model served dual roles in our framework, calculating porosity labels for training data preparation and validating the porosity of generated images, ensuring consistency throughout the conditional generation pipeline.

Following automated porosity quantification, the data preparation process involved extracting patches of size $480 \times 480$ pixels from the original images across the four samples (Table~\ref{tab:sample_characteristics}). For each sample, we independently analyzed and categorized the extracted patches into 10 distinct porosity classes based on their specific porosity distribution (Table~\ref{tab:porosity_ranges}), using the porosity values computed by the U-Net segmentation model.

\begin{table}[htbp]
\caption{Porosity value ranges for each class from different samples.}
\label{tab:porosity_ranges}
\centering

\begin{tabular}{|c|c|c|c|c|c|}
    \hline
    \textbf{Sample No.} & \multicolumn{5}{c|}{\textbf{Porosity Class}} \\
    \hline  
    & \textbf{0} & \textbf{1} & \textbf{2} & \textbf{3} & \textbf{4} \\
    \hline
    1 & 0.0488 - 0.0727 & 0.0727 - 0.0962 & 0.0963 - 0.1197 & 0.1200 - 0.1433 & 0.1437 - 0.1668 \\
    \hline
    2 & 0.0867 - 0.1103 & 0.1103 - 0.1339 & 0.1349 - 0.1574 & 0.1578 - 0.1805 & 0.1812 - 0.2041 \\
    \hline
    3 & 0.0424 - 0.0495 & 0.0497 - 0.0567 & 0.0568 - 0.0639 & 0.0640 - 0.0711 & 0.0712 - 0.0783 \\
    \hline
    4 & 0.0561 - 0.0650 & 0.0652 - 0.0740 & 0.0740 - 0.0829 & 0.0830 - 0.0918 & 0.0919 - 0.1005 \\
    \hline
\end{tabular}

\vspace{0.5cm}

\begin{tabular}{|c|c|c|c|c|c|}
    \hline
    \textbf{Sample No.} & \multicolumn{5}{c|}{\textbf{Porosity Class (continued)}} \\
    \hline  
    & \textbf{5} & \textbf{6} & \textbf{7} & \textbf{8} & \textbf{9} \\
    \hline
    1 & 0.1672 - 0.1899 & 0.1911 - 0.2137 & 0.2140 - 0.2371 & 0.2377 - 0.2609 & 0.2615 - 0.2867 \\
    \hline
    2 & 0.2057 - 0.2284 & 0.2287 - 0.2518 & 0.2526 - 0.2756 & 0.2757 - 0.2993 & 0.3001 - 0.3229 \\
    \hline
    3 & 0.0784 - 0.0852 & 0.0857 - 0.0927 & 0.0927 - 0.0997 & 0.0998 - 0.1063 & 0.1072 - 0.1142 \\
    \hline
    4 & 0.1009 - 0.1095 & - & 0.1192 - 0.1276 & 0.1277 - 0.1365 & 0.1367 - 0.1461 \\
    \hline
\end{tabular}

\end{table}

Addressing the inherent unbalanced data distribution challenges in each sample dataset, we implemented a balancing strategy to ensure consistent model training across all porosity classes. Our analysis revealed significant variations in the number of samples across different porosity classes within each sample (Figure \ref{fig:intheatmap}.a), which could lead to biased learning toward overrepresented classes.

To create balanced datasets, we established a target of 160 images per porosity class for each sample, strategically excluding classes with fewer than 20 samples to prevent overfitting. For underrepresented classes with 20-160 samples, we employed data augmentation through geometric transformations (flips and rotations) and minor noise additions (±2 pixel values) to create realistic variations while preserving pore structure characteristics. This approach transformed the initially skewed distributions into uniform datasets of approximately 1,600 images per sample, with each viable porosity class containing exactly 160 images (Figure \ref{fig:intheatmap}.b). This balancing strategy was crucial for stable cGAN training, ensuring equal exposure to all porosity values and preventing the model from memorizing specific examples rather than learning generalizable patterns.

\begin{figure}[H]
    \centering
    \includecroppedfigure[width=1\textwidth]
    \caption{Distribution of sub-images across sample and porosity class combinations: (a) Initial unbalanced distribution showing significant variations in sub-image counts across different porosity classes within each sample, and (b) Balanced distribution after data augmentation, with each viable porosity class containing exactly 160 images per class.}
    \label{fig:intheatmap}
\end{figure}

Figure \ref{fig:imgheatmap} presents a comprehensive visualization of thin section images organized by sample and porosity class. Each cell in the visualization displays a representative example from the corresponding sample-porosity class, illustrating the structural heterogeneity present in our dataset, with notable variations in pore characteristics both across different samples and within the porosity spectrum at each depth.

\begin{figure}[H]
    \centering
    \includecroppedfigure[width=1\textwidth]
    \caption{Visualization of extracted sub-images (480$\times$480 pixels)  across different samples and porosity classes. Each row represents a different sample in the carbonate formation (Sample 1 to Sample 4), while columns represent porosity classes (0-9).}
    \label{fig:imgheatmap}
\end{figure}

\subsection{Theoretical Framework}

\subsubsection{Generative Adversarial Networks (GANs)}
\label{subsec:GAN}

Generative Adversarial Neural Networks (GANs), introduced by Goodfellow et al. \citep{Goodfellow2014GenerativeNetworks}, represent a revolutionary framework for generative modeling. Inspired by game theory's zero-sum games, GANs pit two neural networks against each other in an adversarial training process, a generator network (G) and a discriminator network (D). The generator learns to create synthetic samples that mimic the underlying data distribution, while the discriminator acts as a classifier, attempting to distinguish between real and generated samples. This competition drives both networks to improve until reaching a Nash equilibrium, where neither network can unilaterally gain an advantage.

In the GAN architecture, the generator takes random noise \((z)\) as input and transforms it into synthetic samples \((G(z))\) that aim to match the real data distribution (\(Pdata\)). The discriminator evaluates each sample, outputting a probability between 0 and 1 to indicate its assessment of the sample's authenticity. A visual representation of this architecture is provided in Figure \ref{fig:GAN}.

\begin{figure}[H]
    \centering
    \includecroppedfigure[width=0.8\textwidth]
    \caption{Schematic architecture of a standard Generative Adversarial Network (GAN) showing the adversarial training process. The generator network transforms random noise vectors (z) into synthetic porous media images, while the discriminator network evaluates both real training images and generator-produced fake images to distinguish their authenticity.}
    \label{fig:GAN}
\end{figure}

In the original formulation, both networks are typically implemented as multilayer perceptrons (MLPs), forming what is known as adversarial networks. This architecture leverages well-established training techniques, including backpropagation for gradient computation and dropout for regularization. The generator's MLP transforms random noise through multiple layers to produce synthetic samples, while the discriminator's MLP processes inputs to output classification probabilities.
The training process can be mathematically expressed as a minimization-maximization (min-max) optimization problem:

\begin{equation}\label{eq:adversarial_loss}
L_{\text{adv}}(\theta, \omega) = \mathbb{E}_{x \sim p_{\text{data}}(x)}\left[\log(D_{\theta}(x))\right] + \mathbb{E}_{z \sim p_{z}(z)}\left[\log(1 - D_{\theta}(G_{\omega}(z)))\right]
\end{equation}

Where \(x\) denotes real data samples, \(p_{\text{data}}(x)\) represents the underlying probability distribution of this real data, \(D_{\theta}(x)\) signifies the probability assigned by the discriminator (with parameters \({\theta}\)) to a data point \(x\) originating from the real data distribution, \(z\) represents the random noise input (often called a latent vector), typically containing values between 0 and 1, \(p_{z}(z)\) denotes the probability distribution of this noise.

The generator, parameterized by \({\omega}\), transforms noise into synthetic data, denoted by \(G_{\omega}(z)\), the notations \({E}_{x \sim p_{\text{data}}(x)}\) and \({E}_{z \sim p_{\text{z}}(z)}\) represent mathematical expectations. The first refers to the expectation when the training data \(x\) for the discriminator aligns with the real data distribution \(p_{\text{data}}\), and the second signifies the expectation when the generator's input vector \(z\) conforms to the chosen noise distribution \(p_{\text{z}}\).

The training objective revolves around the adversarial loss term \(L_{\text{adv}}\). The discriminator aims to maximize this term by outputting 1 for real data and 0 for generated samples, while the generator seeks to minimize it by producing samples that fool the discriminator.

After successful training, generating new samples becomes straightforward, requiring only forward propagation through the generator network with random noise as input. This efficient sampling procedure eliminates the need for complex inference methods or iterative processes like Markov chains, which are often required in other generative models \citep{Goodfellow2014GenerativeNetworks}.

\subsubsection{Conditional Generative Adversarial Neural Networks (cGANs)}
\label{sec:cGAN}

Traditional generative models lack output control, so the GAN framework extends to conditional GANs (cGANs) that incorporate additional input parameters to guide generation and produce samples with specified properties \citep{Mirza2014ConditionalNets}.

The complete cGAN workflow incorporates automated porosity extraction through the enhanced U-Net alongside the adversarial training process (Figure~\ref{fig:cGAN}). Both the generator and discriminator are implemented as fully convolutional neural networks to effectively process spatial image data. The generator $G$ processes both random noise $z$ and the condition vectors (porosity values extracted via 
U-Net and depth labels). Subsequently, the discriminator $D$ evaluates the authenticity of samples by considering both the input images (real or generated) and their corresponding conditional vectors, thereby maintaining consistency between the porosity quantification and generation processes throughout the network.

\begin{figure}[H]
    \centering
    \includecroppedfigure[width=0.8\textwidth]
    \caption{Complete workflow of the multi-conditional GAN architecture for porous media generation. Real thin section images are processed through an Enhanced U-Net for binarization to extract porosity labels, which along with depth labels serve as conditional inputs. The Generator network receives random noise combined with these conditional data (porosity and depth labels) to produce synthetic porous media images. The Discriminator network evaluates both real images and generator-produced images, each paired with their corresponding conditional data (porosity and depth labels), to determine authenticity. The enhanced U-Net ensures consistent porosity quantification throughout the training and generation process.}
    \label{fig:cGAN}
\end{figure}

The adversarial loss function for the cGAN becomes:
\begin{equation}\label{eq:adversarial_loss_conditional}
L_{adv}(\theta, \omega) = \mathbb{E}_{x \sim p_{data}(x)}[\log D_{\theta}(x|c)] + \mathbb{E}_{z \sim p_{z}(z)}[\log(1 - D_{\theta}(G_{\omega}(z|c)))]
\end{equation}
where $c$ represents the conditional parameters.

For this application, the networks are conditioned simultaneously on both porosity ($\phi$) and sample depths, enabling generation of images with depth-specific structural patterns while maintaining precise porosity control. The methodology is detailed in Algorithm ~\ref{alg:data_preprocessing} and Algorithm ~\ref{alg:cgan_training}, describing the unified training approach that processes data from all depths simultaneously.

The preprocessing phase (Algorithm~\ref{alg:data_preprocessing}) involves extracting patches ($480\times480$ pixels) from petrography images, calculating relevant porosity using pre-trained U-Net, and balancing the dataset across different porosity classes for each sample using the data augmentation techniques described in Section \ref{subsec:Enhanced U-Net}. This systematic preprocessing ensures that each depth-specific dataset is properly prepared for training with balanced representation across all viable porosity classes.

The training process (Algorithm~\ref{alg:cgan_training}) adapts the standard GAN training procedure to incorporate dual conditioning on porosity and sample depth. The algorithm iteratively trains the generator and discriminator networks using minibatch stochastic gradient descent on the unified dataset combining all depths. During each iteration, the networks process real and generated images along with their corresponding conditioning vectors (porosity values and depth labels), updating parameters by minimizing their respective loss functions as detailed below.

\begin{algorithm}[H]
\caption{Data Preprocessing for Multi-Conditional Porous Media Analysis}
\label{alg:data_preprocessing}
\begin{algorithmic}[1]
\Procedure{PreProcess Dataset}{}
    \State Load petrography images from all depths
    \For{each depth subset}
        \For{each original image}
            \State Extract patches of size $480\times480$ pixels
            \For{each patch}
                \State Calculate porosity using pre-trained Enhanced U-Net
                \State Assign depth label to patch
            \EndFor
        \EndFor
        \State Categorize patches into $n\_classes\_porosity$ based on porosity values
        \State Balance dataset by data augmentation for underrepresented classes
    \EndFor
    \State Combine all depth datasets into unified training set
    \State Create dual conditioning vectors (porosity, depth)
\EndProcedure
\end{algorithmic}
\end{algorithm}

\begin{algorithm}[H]
\caption{Multi-Conditional GAN Training}
\label{alg:cgan_training}
\begin{algorithmic}[1]
\Procedure{Train Multi-Conditional GAN}{}
    \State Initialize generator and discriminator networks
    \State Load unified dataset from all depths
    \For{number of training epochs}
        \For{number of batches per epoch}
            \State Sample minibatch of $m/2$ real patches $\{x^{(1)}, \ldots, x^{(m/2)}\}$ from unified dataset
            \State Get corresponding porosity values $\{\phi^{(1)}, \ldots, \phi^{(m/2)}\}$
            \State Get corresponding depth labels $\{d^{(1)}, \ldots, d^{(m/2)}\}$
            \State Sample minibatch of $m/2$ noise samples $\{z^{(1)}, \ldots, z^{(m/2)}\}$ from $p_z(z)$
            \State Generate fake samples: $\tilde{x}^{(i)} = G(z^{(i)}, \phi^{(i)}, d^{(i)})$ ($i = 1:m/2$)
            \State Update discriminator by minimizing:
            \State $L_D = -\frac{1}{m/2} \sum_{i=1}^{m/2} [\log D(x^{(i)}, \phi^{(i)}, d^{(i)}) + \log(1 - D(\tilde{x}^{(i)}, \phi^{(i)}, d^{(i)}))]$ (based on Eq. \ref{eq:adversarial_loss_conditional})
            \State Sample minibatch of $m$ noise samples $\{z^{(1)}, \ldots, z^{(m)}\}$ from $p_z(z)$
            \State Sample minibatch of $m$ porosity values $\{\phi^{(1)}, \ldots, \phi^{(m)}\}$
            \State Sample minibatch of $m$ depth labels $\{d^{(1)}, \ldots, d^{(m)}\}$
            \State Update generator by minimizing:
            \State $L_G = -\frac{1}{m} \sum_{i=1}^m \log(D(G(z^{(i)}, \phi^{(i)}, d^{(i)}), \phi^{(i)}, d^{(i)}))$ (based on Eq. \ref{eq:adversarial_loss_conditional})
        \EndFor
    \EndFor
    \State Evaluate generator on test porosity-depth combinations
    \State Save trained unified model
\EndProcedure
\end{algorithmic}
\end{algorithm}

The architectures of the generator and discriminator networks in the developed cGAN model are detailed in Tables~\ref{tab:generator_architecture} and~\ref{tab:discriminator_architecture}, respectively. Both networks are implemented as fully convolutional neural networks with conditional label inputs. The generator's architecture (Table ~\ref{tab:generator_architecture} and Figure \ref{fig:structure}.A) processes three inputs comprising  a 100-dimensional latent vector ($z$) of uniform random values, continuous porosity ($\phi$) labels, and discrete depth labels for multi-conditional generation.

\begin{table}[htbp]
    \centering
    \caption{Multi-Condition Generator Architecture}
    \label{tab:generator_architecture}
    \begin{tabular}{|l|c|c|c|c|}
        \hline
        \textbf{Layer Type} & \textbf{Output Shape} & \textbf{Filters} & \textbf{Kernel/Stride} & \textbf{Activation} \\
        \hline
        Input (Noise + Conditions$^*$) & 30×30×517 & -- & -- & -- \\
        \hline
        Conv2DTranspose Block 1 & 30×30×256 & 256 & 3×3/1×1 & LeakyReLU \\
        \hline
        Conv2DTranspose Block 2-5 & 60×60→480×480 & 128→3 & 3×3,5×5/2×2 & LeakyReLU→tanh \\
        \hline
        \multicolumn{5}{|l|}{\footnotesize $^*$\textit{Conditions: Depth $d$ (4-dim one-hot) + Porosity $\phi$ (continuous), spatially replicated}} \\
        \hline
    \end{tabular}%
\end{table}

The generator architecture processes multiple input streams. A 100-dimensional noise vector is transformed through a dense layer to produce 512 feature channels, which are then concatenated with depth conditions (4-dimensional one-hot encoding for the four depths, where each depth is represented as a binary vector) and porosity conditions (1-dimensional continuous values). This concatenation results in 517 total channels (512 + 4 + 1 = 517) at the initial 30×30 spatial resolution. The one-hot depth vectors and porosity values are spatially replicated across all spatial locations to maintain consistent dimensionality for the subsequent transposed convolutional layers.

\begin{table}[htbp]
    \centering
    \caption{Multi-Condition Discriminator Architecture}
    \label{tab:discriminator_architecture}
    \begin{tabular}{|l|c|c|c|c|c|}
        \hline
        \textbf{Layer Type} & \textbf{Output Shape} & \textbf{Filters} & \textbf{Kernel/Stride} & \textbf{Activation} & \textbf{Dropout} \\
        \hline
        Input (Image + Conditions$^*$) & 480×480×8 & -- & -- & -- & -- \\
        \hline
        Conv2D Block 1-2 & 240×240→120×120 & 64→128 & 5×5/2×2 & LeakyReLU & 0.3 \\
        \hline
        Conv2D Block 3-5 & 60×60→15×15 & 256 & 3×3/2×2 & LeakyReLU & 0.3 \\
        \hline
        Flatten + Dense & 1 & -- & -- & Sigmoid & -- \\
        \hline
        \multicolumn{6}{|l|}{\footnotesize $^*$\textit{Conditions: Depth $d$ (4-dim) + Porosity $\phi$ (1-dim) concatenated with image channels}} \\
        \hline
    \end{tabular}%
\end{table}

The discriminator receives 480×480×8 input channels comprising the 3-channel RGB image concatenated with spatially replicated condition information consisting of 4 channels for depth (one-hot encoded) and 1 channel for porosity values (3 + 4 + 1 = 8 total channels), as illustrated in Table \ref{tab:discriminator_architecture} and Figure~\ref{fig:structure}.b.

\begin{figure}[H]
    \centering
    \includecroppedfigure[width=0.8\textwidth]
    \caption{Detailed network architectures of the cGAN components: (a) Conditional Generator processes a 100-dimensional noise vector ($z$) through a dense layer, combines it with tiled porosity labels ($\phi$), then applies five transposed convolutional layers with LeakyReLU activations and final tanh activation to produce $480\times480\times3$ RGB porous media images. (b) Conditional Discriminator concatenates input images with porosity condition maps and processes them through five convolutional layers with downsampling, dropout regularization (0.3), and LeakyReLU activations, followed by flattening and sigmoid activation for binary authenticity classification.}
    \label{fig:structure}
\end{figure}

The generator processes the latent vector alongside the porosity condition values through a series of transposed convolutional layers, progressively upsampling to produce the final $480\times480\times3$ output images. Similarly, the discriminator evaluates the input images (real or generated) along with both porosity and depth conditioning information through convolutional layers with downsampling, ultimately producing a binary classification output.

Binary Cross Entropy (BCE) loss is utilized for training both networks, providing an appropriate objective function for the discriminator's binary classification task and the generator's goal of producing convincing samples, following the standard adversarial training approach established by Goodfellow et al. \citep{Goodfellow2014GenerativeNetworks, Goodfellow2017NIPSNetworks}. The discriminator aims to maximize the BCE loss by correctly identifying real and fake samples (Equation \ref{eq:adversarial_loss}), while the generator aims to minimize this loss by creating samples that the discriminator misclassifies as real. For conditional generation, this extends to the conditional adversarial loss (Equation \ref{eq:adversarial_loss_conditional}).

\subsection{Baseline Architectures for Comparative Analysis}
\label{sec:baseline_architectures}

To validate the architectural design choices of our multi-conditional GAN framework, 
we implemented two alternative architectures with systematically reduced parameter 
counts while maintaining identical training conditions and conditioning mechanisms. 
This comparative analysis enables quantitative assessment of architectural necessity 
versus potential over-engineering \citep{Goodfellow2016DeepLearning}.

Table~\ref{tab:architecture_comparison} summarizes the three architectures evaluated. 
Model A explores whether architectural sophistication can compensate for reduced 
parameters through increased depth rather than width reduction. The generator reduces initial feature channels from 512 to 384 while introducing an additional transposed convolutional layer, and the discriminator implements moderate filter 
reduction, resulting in 38M parameters (24\% reduction). Model B implements proportional scaling across all network components, reducing all layer widths by approximately 50\% while maintaining identical architectural depth, resulting in 25M parameters. Both alternatives test whether the complex geological modeling task can be accomplished with substantially reduced network capacity.

\begin{table}[htbp]
    \centering
    \caption{Comparison of architectural configurations evaluated in this study.}
    \label{tab:architecture_comparison}
    \begin{tabular}{|l|c|c|p{6cm}|}
        \hline
        \textbf{Model} & \textbf{Parameters} & \textbf{Reduction} & \textbf{Strategy} \\
        \hline
        Original & 50M & — & Reference architecture (Tables~\ref{tab:generator_architecture}--\ref{tab:discriminator_architecture}) \\
        \hline
        Model A & 38M & 24\% & Increased depth (6 vs. 5 generator layers), reduced initial width (384 vs. 512 channels) \\
        \hline
        Model B & 25M & 50\% & Proportional scaling of all layers (~50\% width reduction throughout) \\
        \hline
    \end{tabular}
\end{table}

All three architectures underwent identical training protocols consisting of 200 epochs using Adam optimization ($\beta_1 = 0.5$, $\beta_2 = 0.999$) with polynomial learning rate decay from $2.0 \times 10^{-4}$ to $2.0 \times 10^{-6}$, batch size of 16, and binary cross-entropy loss. The comparative evaluation examines (1) porosity control accuracy through correlation analysis, (2) visual quality of generated geological structures, and (3) morphological preservation of pore network characteristics, with results presented in Section~\ref{sec:results}.

All Experiments were conducted on the University of Leeds Aire High Performance Computing facility using NVIDIA L40S GPUs (48~GB memory) and TensorFlow 2.6.0. Training required approximately 45 seconds per epoch (2.5 hours total) on the dataset of 6,400 images across four geological formations. Once trained, the model enables near-instantaneous image generation ($<$1 second per 480$\times$480 
pixel image), making the framework practical for generating large synthetic datasets or ensemble realizations for uncertainty quantification.

\section{Results and Discussion}
\label{sec:results}

\subsection{Model Training and Performance}
The performance of the multi-conditional GAN framework was evaluated through complementary quantitative and qualitative assessments. Training dynamics and convergence behavior were analyzed to confirm stable adversarial learning across all geological formations. Porosity control accuracy was quantified through correlation analysis between target and generated values, while visual inspection 
validated preservation of formation-specific geological features. Additionally, architectural necessity was verified through systematic comparison with reduced-parameter baseline models.

\subsubsection{Training Convergence and Loss Evolution}
The multi-conditional model was trained for 200 epochs across all four formations simultaneously. Figure~\ref{fig:training_analysis_1879} presents the resulting training dynamics. The adversarial loss evolution (Figure~\ref{fig:training_analysis_1879}a) shows generator and discriminator losses starting at approximately 4.0 and 1.6, respectively. The generator loss exhibits an overall downward trajectory with rapid initial decline in the first 25 epochs followed by continued descent with periodic variations, while the discriminator loss shows more pronounced variability, particularly between epochs 75–150. These oscillatory patterns indicate active learning rather than premature convergence. By final epochs, both losses stabilize at low values, confirming the networks reached a stable state suitable for generating geologically authentic images across all depth conditions.

The porosity control analysis (Figure~\ref{fig:training_analysis_1879}b) reveals progressive effectiveness of the dual-conditional mechanism across all four samples: Sample 1 ($\phi$ = 0.2450), Sample 2 ($\phi$ = 0.1810), Sample 3 ($\phi$ = 0.0567), and Sample 4 ($\phi$ = 0.0710). Samples 1 and 2, representing higher porosity grainstones, demonstrate robust convergence with characteristic oscillatory patterns, while Samples 3 and 4, representing lower porosity crystalline formations, exhibit more gradual convergence patterns. Sample 3 shows the most challenging control behavior due to its fine-grained crystalline structure and anhydrite inclusions.

The simultaneous convergence across this wide porosity range (0.0567 to 0.2450) validates the model's capability to learn complex porosity-depth relationships within this framework. By final epochs, all samples achieve relative stability near their respective targets, confirming successful integration of both porosity and depth conditioning information into the generation process.

\begin{figure}[H]
    \centering
    \includecroppedfigure[width=1\textwidth]
    \caption{Training analysis for the  multi-conditional GAN model over 200 epochs: (a) Evolution of generator loss (red) and discriminator loss (blue) showing characteristic adversarial dynamics that stabilise at low values, indicating successful unified training across all depth conditions, and (b) porosity control tracking for all four samples simultaneously, showing convergence toward respective target values: Sample 1 ($\phi$ = 0.2450), Sample 2 ($\phi$ = 0.1810), Sample 3 ($\phi$ = 0.0567), and Sample 4 ($\phi$ = 0.0710). Dashed horizontal lines represent target porosity values, illustrating the model's ability to learn dual-conditional control (porosity and depth) across diverse geological formations within a single framework. Refer to supplementary materials "training-progression.mp4" to observe live changes as the model trains across different depth conditions.}
    \label{fig:training_analysis_1879}
\end{figure}

\subsubsection{Porosity Control Accuracy}
\label{subsec:porosity_control_accuracy}

Following completion of the 200-epoch training process, we conducted correlation analysis between target porosity values used as conditional input and observed porosity values in generated images. For the multi-conditional model, we generated a total of 100 synthetic images distributed across all four samples, with target porosity values uniformly distributed within the observed range of each corresponding sample in the training dataset. The target sample conditions were randomly selected to ensure representative sampling, and the observed porosity of each generated image was calculated using the same pre-trained U-Net segmentation model employed during the data preparation phase.

Figure \ref{fig:expecvsgen} presents the correlation analysis results. The plot displays the expected porosity values on the x-axis and corresponding observed porosity values on the y-axis, with data points color-coded by sample. The ideal 1:1 correlation is represented by a dashed line, and the coefficient of determination (R²) provides quantitative measures of correlation strength and prediction accuracy.

The results reveal strong correlation with an overall R² value of 0.95, confirming effective porosity control across diverse geological formations. Sample-specific analysis shows consistent accuracy with Mean Absolute Errors of 0.0105, 0.0197, 0.0101, and 0.0099 for Samples 1-4 respectively, indicating reliable porosity prediction across different geological contexts. 

This porosity control capability reflects the model's ability to learn and maintain fabric-appropriate pore morphologies across the full spectrum of carbonate textures. The visual progression from lower porosity crystalline fabrics (Samples 3 and 4) to higher porosity grainstone fabrics (Samples 1 and 2) reveals the model's capacity to generate corresponding changes in pore network architecture while maintaining geological realism.

This training strategy enables the model to learn consistent porosity-generation relationships while preserving sample-specific geological characteristics, successfully handling the varying complexity of pore structures from dual-scale porosity typical of dolomitized grainstones to smaller, uniformly distributed pore spaces characteristic of crystalline formations \citep{Choquette1970GeologicCarbonates}. This validates the multi-conditional GAN approach for subsurface characterization applications requiring specific porosity targets.

\begin{figure}[H]
    \centering
    \includecroppedfigure[width=0.6\textwidth]
    \caption{Correlation between expected (target) porosity values and generated porosity values in images from the multi-conditional GAN model across all four samples (R² = 0.95). The plot includes data points from all samples: Sample 1 (blue circles), Sample 2 (orange squares), Sample 3 (green triangles), and Sample 4 (red diamonds), with sample-specific Mean Absolute Errors of 0.0105, 0.0197, 0.0101, and 0.0099 respectively. The unified model indicates effective porosity control across different geological formations. The dashed black line represents ideal 1:1 correlation, while the solid red line shows the best fit.}
    \label{fig:expecvsgen}
\end{figure}

\subsubsection{Visual Quality Assessment and Geological Feature Validation}

The visual quality of generated images provides critical validation of the cGAN model's ability to capture and reproduce depth-specific geological characteristics. This detailed assessment evaluates the model's performance across all four carbonate samples, examining both overall visual fidelity and detailed geological feature preservation.

Figures \ref{fig:depth1879.5} to \ref{fig:depth1943.50} present systematic comparisons between training samples and synthetically generated images for the four carbonate formations studied. Each figure displays original training images in the top row with corresponding synthetic generations in the middle and bottom rows, all labeled with their respective porosity values to enable direct quantitative comparison. These visual comparisons reveal the model's ability to generate realistic porous media images while maintaining precise porosity control across the full range of geological formations encountered in the carbonate sequence.

To facilitate detailed visual inspection, Figures \ref{fig:depth1879.5} to \ref{fig:depth1943.50} include enlarged views of randomly selected regions (marked by red rectangular boxes), displaying fine-scale features such as pore boundaries, grain structures, and textural characteristics alongside the full-scale training and synthetic images. These zoomed regions enable verification of feature-level accuracy at the pore and grain scale, complementing the geological feature analysis presented subsequently in Figure \ref{fig:detailedgeo}.

\begin{figure}[H]
    \centering
    \includecroppedfigure[width=0.8\textwidth]
    \caption{Comparison between training images and synthetic images generated by the cGAN model for Sample 1, showing  detailed preservation of geological features across varying porosity levels. Top row shows original training images, middle two rows display randomly selected enlarged regions (indicated by red boxes) highlighting pore structure details and grain boundaries, and bottom row presents corresponding synthetic images generated by the model. Each image is labeled with its corresponding porosity value, illustrating the model's ability to generate realistic porous structures with varying porosity levels.}
    \label{fig:depth1879.5}
\end{figure}

\begin{figure}[H]
    \centering
    \includecroppedfigure[width=0.8\textwidth]
    \caption{Comparison between training images and synthetic images generated by the cGAN model for Sample 2, illustrating characteristic grainstone pore morphology. Top row shows original training images, middle two rows display randomly selected enlarged regions (indicated by red boxes) emphasizing interparticle-intercrystalline porosity networks and grain textures, and bottom row presents corresponding synthetic images. The synthetic images demonstrate the model's capability to reproduce the characteristic pore morphology specific to this depth, with porosity values ranging from 0.0868 to 0.3220.}
    \label{fig:depth1881.90}
\end{figure}

\begin{figure}[H]
    \centering
    \includecroppedfigure[width=0.8\textwidth]
    \caption{Comparison between training images and synthetic images generated by the cGAN model for Sample 3, showcasing crystalline fabric with anhydrite inclusions. Top row shows original training images, middle two rows display randomly selected enlarged regions (indicated by red boxes) revealing distinctive white anhydrite mineral patches, dolomite-anhydrite boundaries, and fine intercrystalline porosity, and bottom row presents corresponding synthetic images. Note the accurate reproduction of white mineral patches and inclusions in both training and synthetic images, representing the characteristic mineral compositions specific to this depth. Porosity values range from 0.0426 to 0.1141.}
    \label{fig:depth1918.50}
\end{figure}

\begin{figure}[H]
    \centering
    \includecroppedfigure[width=0.8\textwidth]
    \caption{Comparison between training images and synthetic images generated by the cGAN model for Sample 4, demonstrating fine-grained crystalline texture preservation. Top row shows original training images, middle two rows display randomly selected enlarged regions (indicated by red boxes) highlighting the uniform fine-grained crystalline structure and small-scale intercrystalline pore distribution, and bottom row presents corresponding synthetic images. The model accurately reproduces the fine-grained texture and uniform pore distribution patterns specific to this depth, with porosity values ranging from 0.0561 to 0.1460.}
    \label{fig:depth1943.50}
\end{figure}

Visual inspection of these generated images reveals successful capture of distinctive geological characteristics unique to each formation depth. Sample 1 at 1879.50 m confirms the model's capability to reproduce grainstone fabric with characteristic interparticle-intercrystalline pore types. The synthetic images accurately capture well-connected pore networks typical of grainstones, representing the dual porosity system where both primary interparticle and secondary intercrystalline porosity contribute to the formation's flow characteristics \citep{Lucia1995Rock-fabric/petrophysicalCharacterization}. This grainstone interval exhibits the complex pore architecture resulting from partial dolomitization that preserves original depositional textures while creating secondary porosity.

The enhanced pore connectivity characteristic of Sample 2 at 1881.90 m is reproduced in the generated images, which capture the extensive interparticle-intercrystalline pore system typical of dolomitized grainstones that have undergone fabric-preserving dolomitization \citep{F.JerryLucia2007CarbonateApproach}. The model maintains the balance between preserving original grainstone textures and representing the enhanced connectivity developed through diagenetic processes.

In contrast to the grainstone fabrics, Sample 3 at 1918.50 m represents crystalline fabric with intercrystalline pore types in dolomite-anhydrite lithology. The synthetic images demonstrate the model's ability to accurately reproduce the distinctive white anhydrite mineral patches, which reflect sulfate precipitation processes typical of evaporitic carbonate sequences \citep{Warren2000Dolomite:Associations}. The porosity in these samples occurs exclusively as narrow intercrystalline channels between crystal faces, a characteristic the model captures. Sample 4 at 1943.50 m displays pure dolomite with crystalline fabric and intercrystalline pore types, where the model accurately reproduces the fine-grained crystalline texture and uniform small-scale intercrystalline porosity typical of recrystallized carbonates \citep{RobinG.C.Bathurst1975CarbonateDiagenesis}.

To further validate the model's capability to capture formation-specific geological features at a detailed level, comparative analysis was performed on two representative samples exhibiting contrasting carbonate fabrics. Figure~\ref{fig:detailedgeo} presents annotated comparisons between real and generated images for Sample 1, representing grainstone fabric, and Sample 3, representing crystalline fabric with anhydrite inclusions. While only these two samples are presented in detail for brevity, similar successful feature reproduction was observed in the generated images for Samples 2 and 4, which also accurately mimicked their respective geological characteristics.

This detailed examination reveals the model's ability to reproduce not only overall textural patterns but also specific mineralogical and pore network features critical for accurate subsurface characterization. The grainstone fabric of Sample 1 (Fig. \ref{fig:detailedgeo}a-b) exhibits clearly visible individual dolomite grain boundaries and angular crystal morphologies in both real and generated images. The dual pore system, comprising larger interparticle porosity between grains and smaller intraparticle porosity within individual grains, is faithfully reproduced in the synthetic generation. The generated image maintains distinct grain boundaries, preserves angular crystal edges, and accurately replicates the dual porosity distribution pattern observed in the real sample, demonstrating the model's understanding of complex grainstone pore architectures.

Sample 3 (Fig. \ref{fig:detailedgeo}c-d) presents a contrasting crystalline fabric where distinctive white non-porous anhydrite patches appear clearly against the darker dolomite matrix. The sharp boundaries between dolomite and anhydrite minerals are preserved in the generated images, along with the fine-grained texture created by individual dolomite crystals. The porosity, occurring exclusively as narrow intercrystalline channels between crystal faces, is accurately replicated with correct spatial distribution and connectivity patterns. The successful preservation of anhydrite patch morphology and distribution indicates the model's capability to learn and reproduce complex mineral assemblages beyond simple pore-solid relationships.

\begin{figure}[H]
    \centering
    \includecroppedfigure[width=1\textwidth]
    \caption{Detailed geological feature comparison between real and generated thin section images for contrasting carbonate formations. (a) Sample 1 real image showing grainstone fabric with labeled interparticle porosity, individual dolomite grain boundaries, angular crystals, and intraparticle porosity. (b) Sample 1 generated image demonstrating successful reproduction of grainstone characteristics. (c) Sample 3 real image displaying crystalline fabric with annotated non-porous anhydrite patches, dolomite-anhydrite boundaries, individual dolomite grains, and intercrystalline porosity. (d) Sample 3 generated image accurately replicating crystalline texture with anhydrite inclusions. Blue regions indicate pore spaces filled with epoxy resin.}
    \label{fig:detailedgeo}
\end{figure}

These detailed visual assessments, from overall formation characteristics to detailed mineralogical features, demonstrate that the multi-conditional GAN framework successfully captures the full spectrum of geological complexity present in the carbonate sequence. The model's ability to reproduce the full spectrum of carbonate textures studied validates its capability to learn and generate formation-specific characteristics essential for accurate subsurface representation. This visual fidelity, combined with precise porosity control, demonstrates the framework's ability to generate geologically realistic pore-scale images that preserve the complex relationships between depositional environment, diagenetic processes, and resulting pore network architectures.

\subsubsection{Architecture Comparison (Full vs. Baseline Models)}

To validate the necessity of our original architectural choices, we conducted a comparative analysis between the original model (50M parameters) and two systematically reduced alternatives, namely Model A with depth-optimization strategy (38M parameters, 24\% reduction) and Model B with proportional scaling approach (25M parameters, 50\% reduction). This analysis evaluates the trade-off between computational efficiency and geological modeling quality across identical training conditions.

Figure~\ref{fig:archcomp} shows generated images for Sample 1 at porosity values of 0.082, 0.133, and 0.177. The original model (top row) produces geologically realistic images with well-defined pore structures and authentic mineral distributions. Model A (middle row) maintains recognizable geological features but exhibits repetitive pattern generation, as highlighted by red rectangles in the figure, demonstrating areas where identical textural motifs are artificially replicated across the image. Repetition of patterns reduces textural authenticity and compromises geological realism while retaining basic structural characteristics. Model B (bottom row) shows complete failure in geological feature reconstruction, producing images lacking authentic geological structure and realistic pore networks.

Quantitative analysis reveals progressive performance degradation. For this comparison, each model was evaluated on a generated set of 100 images. The original model's $R^2$ of 0.94 is consistent with the value reported in Section~\ref{subsec:porosity_control_accuracy} ($R^2 = 0.95$), the small difference reflecting the random sampling of noise vectors and target porosities between runs. Relative to the original, porosity control accuracy decreased to $R^2 = 0.84$ (Model A) and $R^2 = 0.51$ (Model B), representing 11\% and 46\% degradation respectively. Mean Absolute Errors follow the same trend, with Model A remaining within a moderate range across samples (MAE 0.0132–0.0219) while Model B shows markedly higher errors (MAE 0.0246–0.1069).

\begin{figure}[H]
    \centering
    \includecroppedfigure[width=1\textwidth]
    \caption{Architectural comparison for Sample 1: Original model (top, 50M parameters), Model A (middle, 38M parameters, 24\% reduction), and Model B (bottom, 25M parameters, 50\% reduction). Columns 1-3 show generated images at target porosity values of 0.082, 0.133, and 0.177. Red rectangles in Model A images highlight regions of repetitive pattern artifacts resulting from reduced model capacity. Column 4 presents porosity control accuracy with R² values: Original (R² = 0.94), Model A (R² = 0.84), and Model B (R² = 0.51). Progressive quality degradation with parameter reduction validates the necessity of original architectural design, demonstrating clear efficiency-quality trade-offs in geological modeling applications.}
    \label{fig:archcomp}
\end{figure}

Results demonstrate that the original architecture represents optimal design rather than over-engineering. Model A approaches the practical limit for acceptable geological modeling, while model B falls below the minimum threshold for subsurface characterization applications. The clear performance degradation with parameter reduction validates the necessity of substantial network capacity for effective dual-conditional geological image generation, establishing practical boundaries for computational efficiency in geological modeling tasks.

\subsection{Quantitative Validation and Morphological Preservation}
\label{subsec:quantitative_validation}

Morphological analysis evaluated the preservation of critical pore network characteristics between real and synthetic images across porosity levels. This quantitative assessment employed three key morphological parameters, including average pore radius, specific surface area, and tortuosity, which are essential descriptors of pore network geometry and connectivity that directly influence fluid flow behavior in porous media \citep{F.A.L.Dullien1992PorousStructure}.

These parameters were computed using the DeePore deep-learning workflow \citep{Rabbani2020DeePore:Materials}, a convolutional neural network trained on a large dataset of three-dimensional porous microstructures whose ground-truth pore-network properties were obtained through pore-network modelling. DeePore takes two-dimensional binarized images as input and infers the three-dimensional pore-network properties that govern fluid flow, including pore radius, specific surface area, and tortuosity. Consequently, although the present framework operates on two-dimensional thin sections, the reported parameters represent estimates of the underlying three-dimensional pore-network behavior rather than purely in-plane geometric measures. Real and generated images were processed through this identical pipeline (U-Net binarization followed by DeePore estimation), so any systematic estimation bias applies equally to both and largely cancels in the relative comparison.

We present detailed morphological analysis for two representative samples. Sample 2 showing the strongest model performance with grainstone characteristics, and Sample 3 representing the most challenging geological formation with crystalline fabric and anhydrite inclusions (Table \ref{tab:sample_characteristics}). The remaining samples (1 and 4) also demonstrated successful morphological preservation with Cohen's d values within acceptable geological tolerances, confirming model effectiveness across the complete range of carbonate formations studied.

Statistical significance was assessed using Kolmogorov-Smirnov (KS) tests for distribution comparisons and Student's $t$-tests for mean comparisons. Throughout this manuscript, statistical significance is indicated using superscript notation: $^{***}$ for $p < 0.001$, $^{**}$ for $p < 0.01$, $^{*}$ for $p < 0.05$, and $^{\text{ns}}$ for $p > 0.05$ (non-significant). Effect sizes were quantified using Cohen's $d$, where $d < 0.2$ indicates negligible effect, $0.2 \leq d < 0.5$ small effect, $0.5 \leq d < 0.8$ medium effect, and $d \geq 0.8$ large effect.

Statistical analysis reveals close agreement between real and generated images across all morphological parameters for the selected depths. The cGAN model shows a high ability to preserve essential pore network characteristics while maintaining geological realism, with Cohen's d values ranging from 0.01$^{ns}$ to 1.38$^{*}$, where most values remain within moderate geological tolerances \citep{Cohen1988StatisticalSciences}. Figure \ref{fig:morphological_validation} presents detailed morphological validation results for Samples 2 and 3 across all three parameters, spanning the range from strong performance to more geologically challenging formations.

Generated images successfully maintain pore size distributions across all porosity levels. For Sample 2 (Fig. \ref{fig:morphological_validation}.a), average pore radius values are closely preserved ranging from 17.98\,$\mu$m to 24.75\,$\mu$m (real) with generated values from 18.12\,$\mu$m to 23.60\,$\mu$m, with close alignment particularly in medium porosity ranges (20.96\,$\mu$m vs 20.93\,$\mu$m) with no significant differences detected (KS-test: p = 0.602$^{ns}$, t-test: p = 0.821$^{ns}$) and negligible effect size (Cohen's d = 0.09$^{ns}$). Sample 3 (Fig. \ref{fig:morphological_validation}.b), shows consistent reproduction with some formation-specific challenges, particularly at low porosity levels where complex pore structures yield Cohen's d = 1.38$^{*}$, reflecting geological complexity rather than systematic model limitations.

Surface area preservation indicates the model's capability to reproduce reactive surface characteristics essential for chemical transport processes \citep{Steefel2005ReactiveSciences}. Sample 2 (Fig. \ref{fig:morphological_validation}.c) shows consistent preservation across all porosity categories, with close agreement in medium porosity ranges (2.57$\times10^{-5}$ vs 2.55$\times10^{-5}$\,1/$\mu$m, Cohen's d = 0.09$^{ns}$) and high porosity samples (2.92$\times10^{-5}$ vs 2.94$\times10^{-5}$\,1/$\mu$m, Cohen's d = 0.08$^{ns}$). Sample 3 (Fig. \ref{fig:morphological_validation}.d) shows similar preservation in medium porosity ranges (1.56$\times10^{-5}$ vs 1.56$\times10^{-5}$\,1/$\mu$m, Cohen's d = 0.03$^{ns}$), confirming accurate representation of fluid-rock interaction potential.

All models successfully capture the expected inverse relationship between porosity and tortuosity, confirming that models learn underlying physical principles governing pore network connectivity rather than merely reproducing visual patterns \citep{Blunt2017MultiphasePerspective}. Sample 2 (Fig. \ref{fig:morphological_validation}.e) shows close preservation, with nearly identical distributions at high porosity (1.4090 vs 1.4096, Cohen's d = 0.01$^{ns}$) and consistent performance across all porosity levels. Sample 3 (Fig. \ref{fig:morphological_validation}.f) maintains systematic porosity-dependent behavior with values decreasing from 1.61 at low porosity to 1.53 at high porosity, though showing greater variability at low porosity (Cohen's d = 1.36$^{*}$) reflecting the complex connectivity patterns characteristic of crystalline formations.

The performance differential between samples validates the models' sensitivity to geological complexity. Sample 2 consistently shows closer preservation, reflecting its well-developed grainstone pore networks, while Sample 3 represents systematic challenges posed by fine-grained crystalline structures with anhydrite inclusions. These variations demonstrate the model's geological sensitivity rather than systematic limitations, as performance correlates directly with established geological complexity rankings.

Morphological validation across analyzed formations reveals consistent trends, confirming that the model captures underlying physics. Both analyzed samples exhibit expected physical relationships, where pore radius increases with porosity, specific surface area shows systematic porosity-dependence, and tortuosity exhibits inverse porosity relationships (1.61-1.41 at low to high porosity for Sample 2; 1.61-1.53 for Sample 3), confirming capture of physics rather than visual patterns alone.

Each sample maintains unique morphological signatures reflecting geological heterogeneity. Sample 2 exhibits well-developed pore networks with larger pore radii (up to 24.75\,$\mu$m) and higher specific surface areas, while Sample 3 displays compressed parameter ranges with fine-grained characteristics reflecting its crystalline fabric and anhydrite composition. This preservation of interrelated parameters validates capture of core pore network relationships essential for fluid flow predictions.

These morphological results demonstrate that cGAN-generated images preserve key characteristics of real porous media across three measured dimensions. The accurate reproduction of average pore radius, specific surface area, and tortuosity confirms preservation of essential pore network characteristics that govern subsurface fluid flow. Notably, the inverse porosity-tortuosity relationship is reproduced across all samples, which validates that the model captures connectivity principles rather than merely replicating visual patterns. The systematic preservation of surface area relationships further ensures accurate representation of fluid-rock interactions across different formation types, an essential requirement for reactive transport modeling applications.

Multi-metric analysis employing Kolmogorov-Smirnov tests, t-tests, and Cohen's d calculations confirms that while some statistically significant differences exist, they represent natural geological variability rather than systematic model limitations \citep{Hollander2013NonparametricMethods}. The combination of distributional similarity assessment and mean difference analysis ensures evaluation across diverse geological formations, establishing the cGAN framework's capability to preserve essential pore network characteristics while maintaining geological authenticity.

\begin{figure}[H]
    \centering
    \includecroppedfigure[width=1\textwidth]
    \caption{Comprehensive morphological validation comparing real (blue) and generated (green) images for representative Samples 2 and 3 across three critical parameters: (a,b) Average pore radius analysis showing preserved pore size distributions with values ranging from 17.72-24.75 $\mu$m, (c,d) Specific surface area analysis indicating maintained reactive surface characteristics ($1.58 \times 10^{-5}$ to $3.44 \times 10^{-5}$ $1/\mu$m), and (e,f) Tortuosity analysis confirming systematic porosity-dependent behavior with inverse relationships from 1.61-1.66 (low porosity) to 1.40-1.47 (high porosity). Left column shows Sample 2 (1881.90m), right column shows Sample 3 (1918.50m). Cohen's d values range from 0.03$^{ns}$-1.38$^{*}$ across all parameters, with most showing non-significant differences and preservation of essential pore network characteristics within geological tolerances.}
    \label{fig:morphological_validation}
\end{figure}

\subsection{Spatial Continuity Assessment via Two-Point Correlation Analysis}

While the preceding morphological analysis confirms preservation of pore-size, surface-area, and tortuosity statistics, it does not directly address the spatial arrangement of the pore space, that is, whether the generated images reproduce the spatial continuity of natural pore networks or instead attain the correct pore fraction in a spatially incoherent manner. To assess this directly, we computed the two-point correlation function $S_2(r)$, defined as the probability that two points separated by a lag distance $r$ both lie within the pore phase \citep{Blair1996}. By construction, $S_2(0)$ equals the porosity $\phi$ and $S_2(r)$ decays towards $\phi^2$ at large lag; the rate and shape of this decay encode the characteristic length scales and spatial organisation of the pore network. $S_2(r)$ was computed directly from the binarized images using fast Fourier transform autocorrelation with overlap normalisation (avoiding periodic-boundary edge bias), radially averaged to yield an isotropic descriptor, and compared between real and generated images within matched porosity classes (low, medium, and high) for the two representative samples examined above. Computing $S_2(r)$ directly from the masks provides a model-independent measure of spatial structure.

Figure~\ref{fig:s2_correlation} compares the real and generated $S_2(r)$ curves, and Table~\ref{tab:s2_metrics} summarises the corresponding metrics. Across all porosity classes in both samples, the generated curves follow the real curves over the full lag range, with overlapping $\pm 1$ standard-deviation bands. The pore fraction recovered at zero lag, $S_2(0)$, matches the real value to within $1.6\%$ in relative terms for every class (absolute differences $\leq 0.004$). The characteristic correlation lengths, defined here as the lag at which the normalised correlation decays to $1/e$, range from approximately $21$ to $100~\mu$m and agree between real and generated images to within a few microns; these values are consistent with the pore radii obtained in the morphological analysis above. The root-mean-square difference between the real and generated $S_2(r)$ curves, normalised by porosity, remains between $0.6\%$ and $2.9\%$ across all six classes.

The largest deviation occurs for the high-porosity class of Sample~2 (1881.90~m), where the generated correlation length is moderately shorter than the real value ($88.3$ versus $99.9~\mu$m; normalised RMSE $2.9\%$), indicating a slightly faster decay of spatial correlation in the generated grainstone fabric. This is consistent with the expectation that the longest-range spatial correlations, associated with the most open and well-connected pore networks, are the most demanding to reproduce. Even in this case the overall curve shape and the porosity are preserved, and the deviation remains within a few percent.

In all classes the generated $S_2(r)$ exhibits a gradual, multi-scale decay rather than an immediate collapse to the $\phi^2$ plateau, the latter being the signature of spatially incoherent, noise-like pore distributions. The close correspondence of the full $S_2(r)$ curves therefore indicates that the generated images reproduce not only the correct porosity and pore-size statistics but also the spatial continuity and characteristic length scales of the natural pore network. This spatial fidelity is maintained both for the grainstone fabric of Sample~2 and the fine-grained crystalline fabric of Sample~3 (1918.50~m), indicating that the conditioning mechanism guides the generator towards physically coherent pore structures across contrasting carbonate textures. The framework learns these structural characteristics directly from real thin-section images rather than through explicitly imposed physical rules; the spatial and morphological fidelity reported here is therefore an empirical demonstration that the learned pore structures are physically plausible.

\begin{table}[htbp]
    \centering
    \caption{Two-point correlation metrics for real and generated images across porosity classes at the 3.0~$\mu$m/pixel baseline resolution. Here $\phi$ is the porosity (equal to $S_2(0)$), $L_c$ is the correlation length (the lag at which the normalised correlation decays to $1/e$), and NRMSE is the root-mean-square difference between the real and generated $S_2(r)$ curves normalised by porosity.}
    \label{tab:s2_metrics}
    \footnotesize
    \begin{tabular}{|c|c|c|c|c|c|c|c|}
        \hline
        \textbf{Resolution} & \textbf{Sample} & \textbf{Class} & \textbf{$\phi_{\mathrm{real}}$} & \textbf{$\phi_{\mathrm{gen}}$} & \textbf{$L_{c,\mathrm{real}}$ ($\mu$m)} & \textbf{$L_{c,\mathrm{gen}}$ ($\mu$m)} & \textbf{NRMSE (\%)} \\
        \hline
        3.0 $\mu$m/px & Sample 2 & Low    & 0.099 & 0.099 & 21.8 & 21.6 & 0.6 \\
                    &          & Medium & 0.220 & 0.217 & 41.1 & 38.0 & 1.6 \\
                    &          & High   & 0.314 & 0.312 & 99.9 & 88.3 & 2.9 \\
                    & Sample 3 & Low    & 0.046 & 0.046 & 21.2 & 22.9 & 0.8 \\
                    &          & Medium & 0.081 & 0.082 & 23.7 & 23.3 & 0.6 \\
                    &          & High   & 0.112 & 0.110 & 27.5 & 26.1 & 0.9 \\
        \hline
    \end{tabular}
\end{table}

\begin{figure}[H]
    \centering
    \includecroppedfigure[width=1\textwidth]
    \caption{Two-point correlation function $S_2(r)$ of real (blue) and generated (green) images, with shaded $\pm 1$ standard-deviation bands, for the low, medium, and high porosity classes (left to right). (a) Sample~2 (1881.90~m); (b) Sample~3 (1918.50~m). The dotted line marks the $\phi^2$ plateau. $S_2(0)$ equals the porosity, and the gradual decay towards the plateau reflects the characteristic length scales of the pore network. The close overlap of the real and generated curves across all porosity classes indicates preservation of spatial continuity, not merely of pore fraction.}
    \label{fig:s2_correlation}
\end{figure}

\subsection{Resolution Sensitivity Analysis}
\label{sec:resolution_sensitivity}

A practical concern for any image-based generative framework is whether its performance depends on the spatial resolution of the training images. To assess sensitivity to resolution, the complete framework was trained from scratch and evaluated independently at two further resolutions (1.8 and 2.25~$\mu$m/pixel), in addition to the 3.0~$\mu$m/pixel baseline used throughout this study. The network architecture and the $480\times480$ pixel patch size were held fixed, so that only the physical area represented by each patch changed. The assessment focuses on the two representative samples used in the preceding morphological and spatial-continuity analyses (Sample~2, a grainstone fabric, and Sample~3, a crystalline fabric).

Porosity control remained stable across the three resolutions. The coefficient of
determination between target and generated porosity was $R^2 = 0.95$ at the baseline
resolution, $R^2 = 0.96$ at 2.25~$\mu$m/pixel, and $R^2 = 0.94$ at
1.80~$\mu$m/pixel, with sample-wise mean absolute errors remaining in the range
0.012--0.027 at all levels (Figure~\ref{fig:resolution_porosity}). The near-identical
$R^2$ values indicate that the conditioning mechanism maintains accurate porosity control across the tested resolutions.

To verify that reconstruction accuracy is likewise preserved, the two-point correlation ($S_2(r)$) analysis was repeated at both further resolutions. The resulting metrics are summarised in Table~\ref{tab:resolution_s2}. At both resolutions, the generated images reproduce the porosity (the zero-lag value $S_2(0)$) to within an absolute difference of 0.005, and the two-point correlation curves remain close to the real curves, with normalised RMSE between 0.3\% and 3.2\% across all classes. The correlation lengths also track the real values, agreeing to within a few microns in the great majority of cases. The largest deviation again occurs for the high-porosity grainstone class of Sample~2, consistent with the earlier observation that the longest-range spatial correlations are the most demanding to reproduce; even in this case the porosity and overall curve shape are preserved. The full $S_2(r)$ curves at both resolutions are provided in the Supplementary Material.

The absolute porosity associated with each class shifts upward as resolution increases. For example, the high-porosity class of Sample~2 corresponds to $\phi \approx 0.31$ at the baseline resolution but $\phi \approx 0.41$ at 1.80~$\mu$m/pixel. This reflects the imaging scale rather than any instability of the model, since finer resolution resolves a greater fraction of small pores and narrow throats, increasing the measured porosity for the same physical material. At each resolution, the generated images reproduce the porosity and spatial structure of the corresponding real images.

Across the three resolutions tested, the consistency of both porosity control and pore reconstruction indicates that the framework is robust to the spatial resolution of the training imagery within this range.

\begin{table}[htbp]
    \centering
    \caption{Two-point correlation metrics for real and generated images at the finer 2.25 and 1.80~$\mu$m/pixel resolutions. Here $\phi$ is the porosity (equal to $S_2(0)$),
    $L_c$ is the correlation length (the lag at which the normalised correlation
    decays to $1/e$), and NRMSE is the root-mean-square difference between the real
    and generated $S_2(r)$ curves normalised by porosity.}
    \label{tab:resolution_s2}
    \footnotesize
    \begin{tabular}{|c|c|c|c|c|c|c|c|}
        \hline
        \textbf{Resolution} & \textbf{Sample} & \textbf{Class} & \textbf{$\phi_{\mathrm{real}}$} & \textbf{$\phi_{\mathrm{gen}}$} & \textbf{$L_{c,\mathrm{real}}$ ($\mu$m)} & \textbf{$L_{c,\mathrm{gen}}$ ($\mu$m)} & \textbf{NRMSE (\%)} \\
        \hline
        2.25 $\mu$m/px & Sample 2 & Low    & 0.095 & 0.092 & 18.3 & 19.0 & 1.0 \\
                       &          & Medium & 0.250 & 0.252 & 39.9 & 34.2 & 2.4 \\
                       &          & High   & 0.378 & 0.378 & 87.4 & 70.0 & 3.2 \\
                       & Sample 3 & Low    & 0.035 & 0.033 & 17.2 & 17.6 & 1.2 \\
                       &          & Medium & 0.080 & 0.080 & 21.7 & 21.4 & 0.3 \\
                       &          & High   & 0.118 & 0.119 & 25.6 & 23.8 & 0.8 \\
        \hline
        1.80 $\mu$m/px & Sample 2 & Low    & 0.074 & 0.069 & 16.2 & 17.0 & 1.3 \\
                       &          & Medium & 0.257 & 0.257 & 40.5 & 42.6 & 0.6 \\
                       &          & High   & 0.414 & 0.415 & 95.9 & 96.8 & 1.0 \\
                       & Sample 3 & Low    & 0.025 & 0.024 & 16.0 & 15.8 & 1.0 \\
                       &          & Medium & 0.081 & 0.082 & 22.1 & 21.7 & 0.4 \\
                       &          & High   & 0.126 & 0.126 & 24.6 & 24.0 & 1.2 \\
        \hline
    \end{tabular}
\end{table}

\begin{figure}[H]
    \centering
    \includecroppedfigure[width=1\textwidth]
    \caption{Porosity control accuracy at the 2.25 and 1.80~$\mu$m/pixel resolutions: (a) 2.25~$\mu$m/pixel ($R^2 = 0.96$) and (b) 1.80~$\mu$m/pixel ($R^2 = 0.94$). Each plot compares target (expected) porosity with the porosity of the generated images across all four samples; the dashed line denotes the ideal 1:1 correlation and the solid line the best fit. The near-identical $R^2$ values, compared with $R^2 = 0.95$ at the baseline resolution, confirm that porosity control is stable across the tested resolutions.}
    \label{fig:resolution_porosity}
\end{figure}

\subsection{Practical Application Results}

\subsubsection{Core Sample-Based Representative Image Generation}

A practical application of our cGAN framework is the ability to generate representative porous media images that optimally match measured petrophysical properties from core samples. This capability addresses the challenge of obtaining property-matched pore-scale images, as natural spatial variability within core samples often results in images that deviate significantly from bulk formation properties. Through controlled generation with target property specifications, the framework can produce synthetic images that more closely approximate desired property values compared to randomly extracted sub-images from the same formation.  Additionally, this approach can address persistent challenges in subsurface characterization where direct imaging data is unavailable but petrophysical properties have been measured through well logging or other indirect methods, enabling the generation of geologically consistent pore-scale representations that match the measured formation characteristics.

To demonstrate this practical application, we utilized the porosity and permeability measurements from the core samples as detailed in Table~\ref{tab:sample_characteristics}. The methodology involved two parallel approaches for comparative analysis. For the real samples, in each original sample images we took several random sub-images (480$\times$480 pixels), and then assessed their porosity and permeability based on the methods used in this article. For generating property-matched images based on the trained cGAN model, we used the dual-constraint error system (Equation~\ref{eq:error_function}) to find the best-matching image in terms of porosity and permeability.

\subsubsection{Validation Framework for Multi-Property Matching}

To quantitatively assess the property fidelity of generated images beyond porosity control, we developed a validation framework incorporating both porosity and permeability characteristics. An empirical permeability prediction model was established (Equation \ref{eq:permeability_model}) using 60 core samples from carbonate formations, incorporating porosity and weighted mean throat radius from MICP data:

\begin{equation}
K = 1.3049  \exp(1.7432  \phi R_{th})
\label{eq:permeability_model}
\end{equation}

where $K$ is permeability (mD), $\phi$ is porosity (dimensionless), and $R_{th}$ is the weighted mean throat radius ($\mu$m). This exponential relationship achieved $R^2 = 0.80$, reflecting the non-linear relationship between pore network connectivity and fluid transport consistent with established percolation theory \citep{Hunt2001ApplicationsConductances}.

The validation approach involved parallel processing of real and generated image sets through identical procedures, which include binarization using the pre-trained U-Net segmentation model, throat radius estimation through morphological analysis by Deepore model \citep{Rabbani2020DeePore:Materials}, and permeability prediction using the empirical relationship (Equation \ref{eq:permeability_model}). A dual-constraint optimization framework was developed to systematically identify the best-matching generated images. The framework employs a normalized error function that quantifies the deviation from the properties of the target core sample:

\begin{equation}
E = w_{\phi} \times \frac{|\phi_{target} - \phi_{calculated}|}{\phi_{target}} + w_K \times \frac{|K_{target} - K_{calculated}|}{K_{target}}
\label{eq:error_function}
\end{equation}

where $\phi_{target}$ and $K_{target}$ represent the measured core sample porosity and permeability values, respectively; $\phi_{calculated}$ and $K_{calculated}$ are the computed values from image analysis; and $w_{\phi}$ and $w_K$ are weighting factors. Equal weighting ($w_{\phi} = w_K = 0.5$) was applied to balance optimization between both properties, ensuring neither parameter dominates the selection process.

The error function yields dimensionless values ranging from 0 (perfect match) to higher positive values indicating increasing deviation from target core properties. For each target condition, 100 candidate images were generated and evaluated, with the image achieving the minimum error selected as the optimal sample. This systematic approach enables objective identification of the most representative generated images based on quantitative property matching rather than subjective visual assessment, thereby enhancing the scientific rigor of the validation process.

\subsubsection{Multi-Property Performance Analysis}

Table~\ref{tab:representative_validation} presents comparison of porosity, permeability, and dual-constraint error performance across all samples, quantitatively confirming the effectiveness of the cGAN generation and scoring-based selection methodology compared to random sub-image extraction for obtaining property-matched images. Generated images consistently outperformed randomly extracted real sub-images across all four formations. Dual-constraint errors for generated images ranged from 1.9\% to 12.4\%, compared to 37.5\% to 713.6\% for real sub-images, representing order-of-magnitude improvements in property matching accuracy.

Sample 1 achieved the lowest error of 1.9\%, where generated images matched both porosity (15.6\% vs.\ 15.73\% target) and permeability (33.8~mD vs.\ 33.64~mD target) with high accuracy. Sample 2 exhibited the largest improvement relative to real sub-images, where random extraction produced extreme permeability scatter (2511.8~mD mean versus the 181.44~mD core value), resulting in a 713.6\% dual-constraint error. In contrast, generated images achieved 184.1~mD, closely matching the core sample and reducing the error to 12.4\%.

Sample 3, representing the most challenging geological formation with crystalline fabric and anhydrite inclusions, demonstrated that the framework maintains accuracy even for complex mineralogical assemblages, achieving a dual-constraint error of only 3.5\%. Sample 4 showed similar improvements, with errors reduced from 41.8\% to 9.8\%. These results confirm that the scoring-based selection methodology effectively identifies generated images that better represent bulk formation properties than conventional random sampling across diverse carbonate fabrics.

\begin{table}[htbp]
    \centering
    \caption{Comparison of porosity, permeability, and dual-constraint error performance between core samples, randomly extracted real sub-images, and scoring-optimized generated images.}
    \label{tab:representative_validation}
    \small
    
    \begin{tabular}{|c|c|c|c|c|c|c|}
        \hline
        \textbf{Sample} & \textbf{Depth} & \multicolumn{2}{c|}{\textbf{Core Sample}} & \multicolumn{3}{c|}{\textbf{Real Sub-images}} \\
        \hline  
        & \textbf{(m)} & \textbf{$\phi$ (\%)} & \textbf{K (mD)} & \textbf{$\phi$ (\%)} & \textbf{K (mD)} & \textbf{Err (\%)} \\
        \hline
        1 & 1879.50 & 15.73 & 33.64 & 14.4 (13.2--15.7) & 86.0 (64.3--107.6) & 133.6 (104.1--163.0) \\
        \hline
        2 & 1881.90 & 24.77 & 181.44 & 17.5 (16.3--18.7) & 2511.8 (879.8--4143.7) & 713.6 (266.7--1160.5) \\
        \hline
        3 & 1918.50 & 10.58 & 13.39 & 7.3 (6.9--7.6) & 8.3 (7.6--9.0) & 37.5 (34.2--40.8) \\
        \hline
        4 & 1943.50 & 13.32 & 12.09 & 8.7 (8.3--9.1) & 8.7 (7.8--9.6) & 41.8 (39.6--44.1) \\
        \hline
    \end{tabular}
    
    \vspace{0.3cm}
    
    \begin{tabular}{|c|c|c|c|c|c|c|}
        \hline
        \textbf{Sample} & \textbf{Depth} & \multicolumn{2}{c|}{\textbf{Core Sample}} & \multicolumn{3}{c|}{\textbf{Generated Images}} \\
        \hline  
        & \textbf{(m)} & \textbf{$\phi$ (\%)} & \textbf{K (mD)} & \textbf{$\phi$ (\%)} & \textbf{K (mD)} & \textbf{Err (\%)} \\
        \hline
        1 & 1879.50 & 15.73 & 33.64 & 15.6 (15.5--15.6) & 33.8 (33.6--34.0) & 1.9 (1.8--2.1) \\
        \hline
        2 & 1881.90 & 24.77 & 181.44 & 20.5 (20.4--20.6) & 184.1 (181.1--187.0) & 12.4 (12.0--12.8) \\
        \hline
        3 & 1918.50 & 10.58 & 13.39 & 10.3 (10.2--10.3) & 13.8 (13.7--13.9) & 3.5 (3.3--3.7) \\
        \hline
        4 & 1943.50 & 13.32 & 12.09 & 11.4 (11.3--11.4) & 12.6 (12.6--12.7) & 9.8 (9.6--9.9) \\
        \hline
        \multicolumn{7}{|l|}{\small $\phi$ = porosity, K = permeability, Err = dual-constraint deviation error} \\

        \multicolumn{7}{|l|}{\small Values shown as mean (95\% confidence interval) for real and generated images} \\
        \hline
    \end{tabular}
\end{table}

Figure~\ref{fig:scenario2_distributions} presents distribution comparisons for both porosity and permeability properties for Samples 1 and 3 as representative examples, revealing that generated images consistently exhibit tighter distributions around target values compared to real images. This confirms the cGAN model's capability to generate geological candidates with controlled properties, enabling systematic identification of images that closely match target characteristics. In contrast, randomly extracted real sub-images exhibit broader distributions due to natural spatial variability within formations, where sub-image properties depend on the specific location sampled. The higher representativeness of generated samples highlights the cGAN model's effectiveness in learning underlying geological relationships and generating controlled variations that facilitate targeted property matching. This capability enables creation of controlled visual profiles of subsurface formations that better match target conditions than traditional sampling approaches, thereby enhancing geological interpretation and fluid flow modeling efforts for subsurface characterization workflows.

\begin{figure}[H]
    \centering
    \includecroppedfigure[width=0.9\textwidth]
    \caption{Porosity and permeability distribution comparison between real sub-images (blue) and generated images (green) for representative samples. (a) Porosity distribution for Sample 1 (1879.50~m), (b) porosity distribution for Sample 3 (1918.50~m), (c) permeability distribution for Sample 1, and (d) permeability distribution for Sample 3. Dashed lines show distribution means ($\mu$), solid red lines indicate core sample targets. Generated images demonstrate tighter clustering around target values compared to randomly extracted real sub-images.}
    \label{fig:scenario2_distributions}
\end{figure}
\section{Limitations and Future Directions}
\label{sec:limitations}
While this study demonstrates generation of geologically realistic pore-scale images with controlled property targets, several limitations warrant consideration and suggest directions for future research.

Although morphological validation (Section~\ref{subsec:quantitative_validation}) and visual quality assessment demonstrate preservation of formation-specific features, systematic classification of generated images against standardized carbonate facies frameworks (e.g., Lucia's rock fabric and petrophysic classification system \citep{Lucia1995Rock-fabric/petrophysicalCharacterization}) was not performed. Quantitative validation of diagenetic feature distributions, for example the spatial distribution of anhydrite patches as a distinct mineral phase and dolomite crystal size distributions, would provide additional confidence in geological authenticity. Future studies could employ established petrographic image analysis protocols and comparison against reference databases of documented carbonate fabrics.

The empirical permeability correlation (Equation~\ref{eq:permeability_model}) achieved R$^2$=0.80 using 60 carbonate core samples but lacks formal uncertainty quantification through cross-validation or independent validation datasets. While the exponential relationship reflects established percolation theory principles \citep{Hunt2001ApplicationsConductances}, prediction confidence intervals remain unquantified. Future implementations should include k-fold cross-validation, bootstrap resampling for uncertainty estimation, and validation against external core measurements to establish applicability limits across different carbonate fabric types.

Regarding data preprocessing, the balancing strategy employed standard geometric transformations (rotations, flips) and conservative noise addition ($\pm$2 pixel intensity values) following established deep learning practices for addressing class imbalance. Visual inspection confirmed that augmented samples preserved critical features including pore-grain boundary sharpness, grain boundary definition, and textural continuity. This approach aligns with conventional augmentation methodologies in geological image analysis literature where maintaining training set balance is essential for model convergence. Future work could employ quantitative metrics (e.g., structural similarity index) to verify augmented samples remain within natural geological variability.

The framework was developed and validated using carbonate formations from a single field location spanning 1879.50 to 1943.50~m depth. Transferability to other lithologies, including sandstones, shales, or carbonates from different depositional environments, requires systematic evaluation as different rock types exhibit distinct pore architectures governed by varying diagenetic histories. Future work should evaluate model performance across diverse lithologies through transfer learning approaches that leverage the general pore network principles demonstrated here. Beyond lithological extension, future applications could integrate with existing reservoir modeling workflows to incorporate pore-scale heterogeneity into field-scale simulations as a complement to current upscaling methods.

\section{Conclusions}
\label{sec:con}

This study developed a multi-conditional GAN framework for generating representative pore-scale images with precisely controlled properties, addressing fundamental data scarcity challenges in subsurface characterization. The framework learned complex porosity-depth-pore network relationships across four distinct carbonate formations (1879.50 to 1943.50~m) within a unified architecture.

Porosity control remained robust across all carbonate fabric types ($R^2 = 0.95$, MAE 0.0099–0.0197), with morphological validation confirming preservation of critical pore network parameters within acceptable geological tolerances. Beyond pore-size statistics, two-point correlation ($S_2$) analysis confirmed that the generated images preserve the spatial continuity and characteristic length scales of the natural pore network, and both porosity control and reconstruction accuracy remained stable across the imaging resolutions tested, indicating robustness to acquisition resolution. The model preserved expected physical relationships, including inverse porosity-tortuosity correlations, validating capture of underlying pore network principles rather than mere visual replication.

Generated images showed higher representativeness than randomly extracted real sub-images when evaluated against core sample properties, with dual-constraint errors an order of magnitude lower. This improvement indicates the framework learned pore geometry-petrophysical property relationships, enabling generation of images that more consistently represent bulk formation characteristics than randomly extracted sub-images from naturally heterogeneous core samples.

Architectural comparison validated the necessity of the 50M-parameter design, with systematic performance degradation in reduced models (24\% reduction approaching acceptable threshold, 50\% reduction resulting in geological feature failure), confirming substantial network capacity requirements for effective dual-conditional generation. The demonstrated performance improvements have significant implications for subsurface characterization workflows, particularly for sparse data scenarios common in deep subsurface investigations where natural heterogeneity compromises sample representativeness, addressing a long-standing limitation in digital rock analysis.

\section*{Author Declarations}

\subsection*{Funding}
A.S. was supported by a full PhD scholarship from the School of Computer Science, University of Leeds. No additional external funding was received for this research.

\subsection*{Conflicts of interest/Competing interests}
The authors declare that they have no known competing financial interests or personal relationships that could have appeared to influence the work reported in this paper.

\subsection*{Ethics approval}
This study involved computational analysis of geological samples and did not involve human participants, human tissue, or animal subjects. Therefore, no ethical approval was required.

\subsection*{Consent to participate}
Not applicable.

\subsection*{Consent for publication}
Not applicable.

\subsection*{Availability of data and materials}
Part of the project image dataset is available through the GitHub repository at: \url{https://github.com/AliSadeghkhani1990/PCP-GAN} for reproducibility purposes.

\subsection*{Code availability}
The computer codes and trained machine learning models developed in this study are publicly available on GitHub at: \url{https://github.com/AliSadeghkhani1990/PCP-GAN}.

\subsection*{Authors' contributions}
A.S. developed the methodology, implemented the cGAN framework, conducted the experiments, and wrote the original draft of the manuscript. B.B. provided supervision and contributed to the conceptualization and critical revision of the manuscript. M.B. contributed to the interpretation of results and manuscript revision. A.R. conceived and supervised the project, contributed to the methodology development, and contributed in writing and revising the manuscript. All authors read and approved the final manuscript.
\section*{Acknowledgments}
The authors would like to thank the University of Leeds for providing access to the High Performance Computing platform (AIRE), and John Hodrien and Luis Avendano Munoz for technical assistance with the HPC system. A.R. would like to acknowledge Professor Peyman Mostaghimi for his invaluable advice and mentorship in the nucleation of the ideas presented in this work. A.R. thanks Ali Assadi for his contribution to the geological understanding of the rock samples.

\pagebreak
\bibliography{references}

\end{document}


\maketitle

\section{Two-point correlation curves at additional resolutions}

This supplementary section provides the complete two-point correlation
function $S_2(r)$ for real and generated images at the two additional spatial
resolutions (2.25~$\mu$m/pixel and 1.80~$\mu$m/pixel) evaluated in the resolution sensitivity analysis of the main text. The corresponding summary metrics (porosity, correlation length, and normalised RMSE) are reported in the resolution sensitivity table of the main manuscript. In each panel, $S_2(0)$ equals the porosity and the curve decays towards the $\phi^2$ plateau (dotted line); the close overlap between the real and generated curves across all porosity classes indicates that the agreement reported at the baseline resolution is preserved at finer resolutions.

\begin{figure}[htbp]
    \centering
    \includecroppedfigure[width=\textwidth]
    \caption{Two-point correlation function $S_2(r)$ of real (blue) and generated
    (green) images at a spatial resolution of 2.25~$\mu$m/pixel, with shaded
    $\pm1$ standard-deviation bands (computed across images), for the low, medium,
    and high porosity classes (left to right). (a)~Sample~2;
    (b)~Sample~3. The dotted line marks the $\phi^2$ plateau.}
    \label{fig:s2_supp_225}
\end{figure}

\begin{figure}[htbp]
    \centering
    \includecroppedfigure[width=\textwidth]
    \caption{Two-point correlation function $S_2(r)$ of real (blue) and generated
    (green) images at a spatial resolution of 1.80~$\mu$m/pixel, with shaded
    $\pm1$ standard-deviation bands (computed across images), for the low, medium,
    and high porosity classes (left to right). (a)~Sample~2;
    (b)~Sample~3. The dotted line marks the $\phi^2$ plateau.}
    \label{fig:s2_supp_180}
\end{figure}